\documentclass{article}

\usepackage[numbers]{natbib}
\usepackage[final]{neurips_2019}

\usepackage[utf8]{inputenc} % allow utf-8 input
\usepackage[T1]{fontenc}    % use 8-bit T1 fonts
\usepackage{hyperref}       % hyperlinks
\usepackage{url}            % simple URL typesetting
\usepackage{booktabs}       % professional-quality tables
\usepackage{amsfonts}       % blackboard math symbols
\usepackage{nicefrac}       % compact symbols for 1/2, etc.
\usepackage{microtype}      % microtypography

\usepackage{graphicx}
\usepackage{pgfplots}
\usepackage{wrapfig}

\usepgfplotslibrary{colorbrewer}
\usepgfplotslibrary{fillbetween}
\pgfplotsset{compat=1.14}
\pgfplotsset{cycle list/Dark2}

\newcommand{\dataset}[1]{\textsc{#1}}

\title{What is needed for simple spatial language capabilities in VQA?}

\author{%
Alexander Kuhnle \\
Department of Computer Science \\
and Technology \\
University of Cambridge \\
{\tt aok25@cam.ac.uk} \\\And
Ann Copestake \\
Department of Computer Science \\
and Technology \\
University of Cambridge \\
{\tt aac10@cam.ac.uk} \\
}

\begin{document}

\maketitle

\begin{abstract}
Visual question answering (VQA) comprises a variety of language capabilities. The diagnostic benchmark dataset CLEVR has fueled progress by helping to better assess and distinguish models in basic abilities like counting, comparing and spatial reasoning \textit{in vitro}.
%However, a range of models now exhibit close-to-perfect performance on this dataset, which makes comparing them difficult.
Following this approach, we focus on spatial language capabilities and investigate the question: what are the key ingredients to handle simple visual-spatial relations? We look at various VQA models and evaluate their learning behavior on diagnostic data which is solely focused on spatial relations.
% We take the SAN, RelNet, FiLM and MC models due to their relative simplicity and similarity to the CNN-LSTM baseline, and evaluate their learning behavior on diagnostic data similar to CLEVR, but which is solely focused on spatial relations.
Via comparative analysis and targeted model modification we identify what really is required to substantially improve upon the CNN-LSTM baseline.
\end{abstract}

\section{Introduction}

Visual question answering is a broad and high-level task formulation which requires a wide range of visual and language understanding abilities. The VQA Dataset \cite{Antol2015} tries to capture the variety of generic real-world visual questions, but while accuracy scores on this dataset have increased over time, it remained unclear what techniques consistently improve performance for which type of questions/images. In reaction to unsatisfying evaluation results, the CLEVR dataset \citep{Johnson2017a} was introduced with the motivation to explicitly focus on a more well-defined subset of diagnostic VQA instances \textit{in vitro}. An important finding was that some of the previous state-of-the-art models did not, in fact, substantially improve upon simple baselines.

In the meantime, various models have been proposed for the CLEVR dataset, many of which reach close-to-perfect performance \citep{Hu2017,Johnson2017b,Santoro2017,Perez2018,Hudson2018,Mascharka2018,Malinowski2018,Yang2018}. Marginal differences between models can be deceptive since, on the one hand, it is unclear what the real strength of each model is and, on the other hand, exactly which part of the model contributed to what observed improvement. Taking into account that CLEVR is a decidedly diagnostic dataset, comparable performance levels rather indicate the limits of differentiation of the dataset, and consequently call for further assessment.

In this paper we focus on simple spatial language understanding. Since CLEVR contains questions with spatial relations like \textit{``left of''} or \textit{``in front of''}, we assume that the recent CLEVR models are able to correctly process such phrases. The question we thus seek to answer is: what architectural module(s) enable a model to handle spatial descriptions? In particular, we investigate four VQA models, all of which follow the architecture pattern of the CNN-LSTM baseline, but introduce modifications which improve model capabilities substantially: SAN with stacked attention \citep{Yang2016}, RelNet with its relation module \citep{Santoro2017}, FiLM with feature-wise linear modulation \citep{Perez2018}, and the multimodal core \citep{Malinowski2018}.

\begin{wrapfigure}{r}{0.5\linewidth}
\begin{center}
\begin{minipage}{0.37\linewidth}
\includegraphics[width=\linewidth]{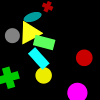}
\end{minipage}%
\hspace{0.03\linewidth}%
\begin{minipage}{0.48\linewidth}
\small
\textbf{Explicit relation:}\par
\textit{A rectangle is closer to the\\ triangle than a red circle.}
\par\vspace{0.2cm}
\textbf{Implicit (comparative):}\par
\textit{The lower cross is green.}
\par\vspace{0.2cm}
\textbf{Implicit (superlative):}\par
\textit{The leftmost circle is gray.}
\end{minipage}
\end{center}
\caption{\label{figure:example}%
An example image plus three valid captions containing simple spatial relations. Each one illustrates a different way of how such a relation may be realized in language. The statements in our experiments may also be wrong, and thus basically act as yes/no questions, which the model has to answer correctly, that is, infer whether or not a statement agrees with an image.}
\end{wrapfigure}

We analyze performance of these models on visually grounded spatial language data produced by the ShapeWorld simulator \citep{Kuhnle2017}, which shares the abstract domain of colored shapes with CLEVR and can be configured to only produce spatial statements of a certain type. Figure \ref{figure:example} shows an example image together with three statements about the image. Note that, technically, the task here is about identifying
image-caption agreement, but the data can be trivially transformed into `pseudo-VQA' data by adding a corresponding yes/no answer.

Our results indicate that two alternative techniques enable models to achieve a high accuracy for VQA instances involving spatial relations: concatenating image features with relative spatial coordinates is easier to integrate with any architecture, whereas feature-wise linear modulation for early fusion of modalities in combination with convolutional layers is the more effective method. Other features, like stacked attention or the relation module of RelNet, did not on its own contribute to improved performance.

\section{Data}

We use the ShapeWorld system \citep{Kuhnle2017} to produce the data for our experiments. Each data point consists of an image of size $64 \times 64$ and shows four to ten colored two-dimensional shapes randomly located on a black background, accompanied by a natural language statement and a binary agreement value, which we interpret as `question' and yes/no answer.

We experiment with three types of statements requiring spatial understanding. In the first, referred to as \dataset{spatial-explicit}, the spatial relation acts as the verb phrase of the sentence: \textit{``X is to the left of Y.''}. For the other two, referred to as \dataset{spatial-implicit}, the relation is realised as adjectival predication, either in its positive/comparative (\dataset{spatial-comparative}) or superlative form (\dataset{spatial-superlative}): \textit{``The left/leftmost X is Y.''}. The following six/eight spatial relations are available in the respective forms: \textit{``left/-most''}, \textit{``right/-most''}, \textit{``above/upper/-most''}, \textit{``be\-low/lower/-most''}, \textit{``closer/-est to the X than''}, \textit{``farther/-est from the X than''}, \textit{``behind''} and \textit{``in front of''}. Of these, the latter two are only generated for \dataset{spatial-explicit}, since the other two variants are unlikely to be found in randomly sampled images. Figure \ref{figure:example} shows an image plus an agreeing example for each caption type. Note that ShapeWorld only produces statements that `make sense`: for instance, \dataset{spatial-comparative} instances only if there are two objects satisfying the noun phrase description, or \dataset{spatial-superlative} if there are at least two.

\section{Models}

Besides the unimodal CNN/LSTM and the multimodal CNN-LSTM baseline, we investigate four recent VQA models: SAN \citep{Yang2016}, RelNet \citep{Santoro2017}, FiLM \citep{Perez2018}, and MC \citep{Malinowski2018}. We identify what the corresponding paper introduces as the core module, and try to keep hyperparameters the same across all models for the other generic parts, to enable a fair comparison of the actual innovation behind the model. Generic architecture parts are the image and language feature extractor as well as the answer classification part. The core module is the part responsible for combining image and language features to arrive at a final pre-answer embedding: the stacked attention layers for SAN, the relation module for RelNet, the linear modulation layers for FiLM, and the multimodal core module for MC\footnote{Model implementations can be found on GitHub under \url{https://github.com/AlexKuhnle/film}.}.

\paragraph{Image module.}
The visual input is processed by a stack of three convolution layers, each with $128$ kernels of size $3 \times 3$ and stride $2$ followed by batch normalization and ReLU activation, thus producing image features of size $8 \times 8 \times 128$.

\vspace{-0.15cm}

\paragraph{Language module.}
The words of the language input are mapped to $128$-dimensional embeddings and subsequently processed by an LSTM, or GRU in case of FiLM, of size $512$, or $128$ in case of RelNet (to keep pairwise combinations small), thus producing a language embedding of size $512$/$128$.

\vspace{-0.15cm}

\paragraph{Core modules.}
The implementations mostly follow the description in the original papers. Minor differences: In case of RelNet, instead of the image module producing 24-dim.\ features, we apply a 32-dim.\  linear layer before the relation module. In case of FiLM, we apply a final 128- as opposed to 512-dim.\  linear layer after the FiLM layers. In case of SAN, we follow the implementation as part of the CLEVR/FiLM GitHub repository, but use 256- as opposed to 512-dim.\  stacked attention layers and also apply an initial 256-dim.\  linear layer to the language input.

\vspace{-0.15cm}

\paragraph{Classification module.}
The core module output is processed by a $1024$-dimensional linear layer followed by batch normalization and ReLU activation, before producing a distribution over answers.

\vspace{-0.15cm}

\paragraph{Optimization.}
Models are optimized using Adam with learning rate $3 \cdot 10^{-4}$ and batch size $64$.

\section{Experimental setup}

\begin{figure*}
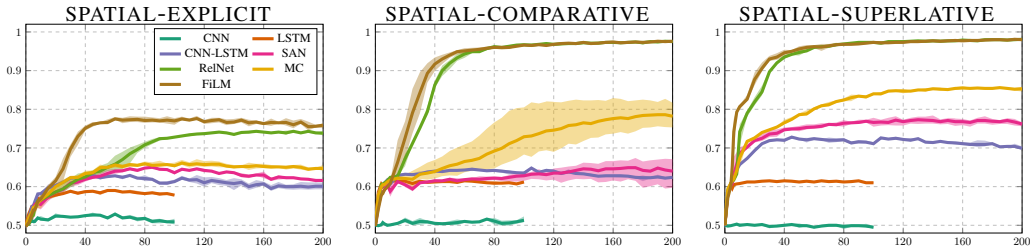


\begin{minipage}{0.32\linewidth}
\centering
\dataset{spatial-explicit}\par
\resizebox{\linewidth}{!}{
\begin{tikzpicture}[baseline=(current bounding box.center)]
\begin{axis}[
every axis plot/.append style={line width=2.5pt},
cycle list name=Dark2,
mark options={mark=none},
width=10cm,
height=7.5cm,
xmin=0, xmax=200,
xtick={0,40,80,120,160,200},
xticklabel style={font=\footnotesize},
xmajorgrids,
ymin=0.48, ymax=1.02,
ytick={0.5,0.6,0.7,0.8,0.9,1.0},
ymajorgrids,
yticklabel style={font=\footnotesize},
grid style=dashed,
legend columns=2
]
\input{spatial-explicit.tex}
]
\end{axis}
\end{tikzpicture}}
\end{minipage}

\begin{minipage}{0.32\linewidth}
\centering
\dataset{spatial-comparative}\par
\resizebox{\linewidth}{!}{
\begin{tikzpicture}[baseline=(current bounding box.center)]
\begin{axis}[
every axis plot/.append style={line width=2.5pt},
cycle list name=Dark2,
mark options={mark=none},
width=10cm,
height=7.5cm,
xmin=0, xmax=200,
xtick={0,40,80,120,160,200},
xticklabel style={font=\footnotesize},
xmajorgrids,
ymin=0.48, ymax=1.02,
ytick={0.5,0.6,0.7,0.8,0.9,1.0},
ymajorgrids,
yticklabel style={font=\footnotesize},
grid style=dashed,
legend columns=2
]
\input{spatial-comparative.tex}
]
\end{axis}
\end{tikzpicture}}
\end{minipage}

\begin{minipage}{0.32\linewidth}
\centering
\dataset{spatial-superlative}\par
\resizebox{\linewidth}{!}{
\begin{tikzpicture}[baseline=(current bounding box.center)]
\begin{axis}[
every axis plot/.append style={line width=2.5pt},
cycle list name=Dark2,
mark options={mark=none},
width=10cm,
height=7.5cm,
xmin=0, xmax=200,
xtick={0,40,80,120,160,200},
xticklabel style={font=\footnotesize},
xmajorgrids,
ymin=0.48, ymax=1.02,
ytick={0.5,0.6,0.7,0.8,0.9,1.0},
ymajorgrids,
yticklabel style={font=\footnotesize},
grid style=dashed,
legend columns=2
]
\input{spatial-superlative.tex}
]
\end{axis}
\end{tikzpicture}}
\end{minipage}

\caption{\label{figure:results1}%
Performance curves over the course of training (x-axis: iterations in 1000, y-axis: accuracy).}
\end{figure*}

We run separate experiments for each of the three caption types, since we noted that performance levels and learning behaviors vary noticeably between them. For each type, 500k training instances and 10k validation instances are generated, following the same generator configuration/distribution as the training data. We run every experiment three times and generally refer to the average accuracy, but additionally indicate minimum/maximum observed performance amongst the three runs in figures as shaded area. Models are trained for 200k iterations (100k in the case of the CNN/LSTM baseline), which corresponds to roughly 25 epochs given the batch size of 64.

\section{Results}

Figure \ref{figure:results1} shows how performance develops over the course of training for each model on the three datasets. A first observation is that no model reaches an accuracy of more than 80\% on the \dataset{spatial-explicit} dataset. While accuracy varies across relations -- \textit{``left/right/above/below''} appear generally easier to learn than the other four relations, particularly \textit{``closer/farther\dots than''} -- the main reason seems to be that the evaluated models struggle with multiple mentions of different shapes, which captions in the other datasets do not contain. Performance on statements like \textit{``A square is above a circle.''} is only around 66\%, whereas statements like \textit{``A red shape is above a blue shape.''} produce the correct response with an accuracy of 95\% in case of the FiLM model, but others show comparable differences. We leave further investigation of this finding to future research, and work on basis of the still substantial gap observed between some models.

The other obvious finding is that, despite their comparable performance on CLEVR, the models under consideration exhibit markedly different learning behavior and final accuracy levels. On the \dataset{spatial-explicit} dataset, FiLM clearly dominates with an eventual accuracy of around 77\% reached after only 50k iterations. RelNet is the only other model which catches up later on and reaches a final performance of 74\%, while the others do not improve by more than 5\% upon the CNN-LSTM baseline. On the two \dataset{spatial-implicit} datasets, FiLM and RelNet show virtually the same learning curve and solve the dataset almost perfectly after around 60-80k iterations, with around 97\% final accuracy. The MC model is second-best with an accuracy of $\sim$78\% for \dataset{spatial-comparative} and 85\% for \dataset{spatial-superlative}, whereas SAN only shows slightly better performance than the CNN-LSTM baseline on \dataset{spatial-superlative}.

\begin{figure*}
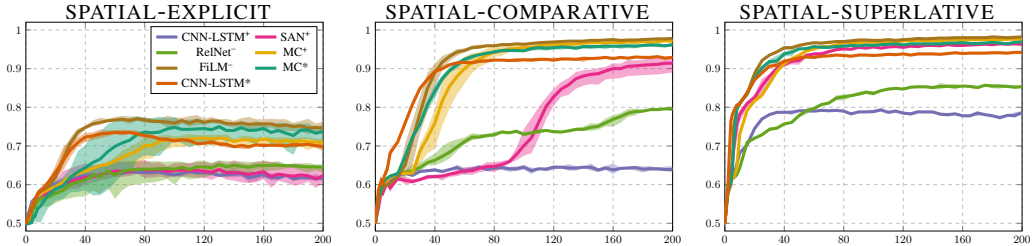


\begin{minipage}{0.32\linewidth}
\centering
\dataset{spatial-explicit}\par
\resizebox{\linewidth}{!}{
\begin{tikzpicture}[baseline=(current bounding box.center)]
\begin{axis}[
every axis plot/.append style={line width=2.5pt},
cycle list name=Dark2,
mark options={mark=none},
width=10cm,
height=7.5cm,
xmin=0, xmax=200,
xtick={0,40,80,120,160,200},
xticklabel style={font=\footnotesize},
xmajorgrids,
ymin=0.48, ymax=1.02,
ytick={0.5,0.6,0.7,0.8,0.9,1.0},
ymajorgrids,
yticklabel style={font=\footnotesize},
grid style=dashed,
legend columns=2
]
\input{spatial-explicit-coords.tex}
]
\end{axis}
\end{tikzpicture}}
\end{minipage}

\begin{minipage}{0.32\linewidth}
\centering
\dataset{spatial-comparative}\par
\resizebox{\linewidth}{!}{
\begin{tikzpicture}[baseline=(current bounding box.center)]
\begin{axis}[
every axis plot/.append style={line width=2.5pt},
cycle list name=Dark2,
mark options={mark=none},
width=10cm,
height=7.5cm,
xmin=0, xmax=200,
xtick={0,40,80,120,160,200},
xticklabel style={font=\footnotesize},
xmajorgrids,
ymin=0.48, ymax=1.02,
ytick={0.5,0.6,0.7,0.8,0.9,1.0},
ymajorgrids,
yticklabel style={font=\footnotesize},
grid style=dashed,
legend columns=2
]
\input{spatial-comparative-coords.tex}
]
\end{axis}
\end{tikzpicture}}
\end{minipage}

\begin{minipage}{0.32\linewidth}
\centering
\dataset{spatial-superlative}\par
\resizebox{\linewidth}{!}{
\begin{tikzpicture}[baseline=(current bounding box.center)]
\begin{axis}[
every axis plot/.append style={line width=2.5pt},
cycle list name=Dark2,
mark options={mark=none},
width=10cm,
height=7.5cm,
xmin=0, xmax=200,
xtick={0,40,80,120,160,200},
xticklabel style={font=\footnotesize},
xmajorgrids,
ymin=0.48, ymax=1.02,
ytick={0.5,0.6,0.7,0.8,0.9,1.0},
ymajorgrids,
yticklabel style={font=\footnotesize},
grid style=dashed,
legend columns=2
]
\input{spatial-superlative-coords.tex}
]
\end{axis}
\end{tikzpicture}}
\end{minipage}

\caption{\label{figure:results2}%
Performance curves, for models with (\textsuperscript{+}) or without (\textsuperscript{--}) coordinate map, or with early FiLM fusion and convolutions instead of concatenation and fully-connected layers (*).}
\end{figure*}

What is the reason for the superior performance of FiLM and RelNet? We conjectured that it may be due to the fact that these two are the only models that attach a map of relative spatial coordinates to the image features. This minor detail infuses useful spatial information and thus relieves the core module of having to learn the concept of relative spatial position from scratch. Since the modification can easily be applied to the other models (at the beginning of their core module), we run the same experiments with added coordinate information for MC, SAN and CNN-LSTM, and instead remove it from FiLM and RelNet. Results are shown in figure \ref{figure:results2}.

Comparing the MC and RelNet model, MC with or without coordinates performs roughly on par with the RelNet model under the same condition. Since their architectures mainly differ in whether to process single positional image embeddings or pairwise concatenations thereof, this result indicates that the relation module, which is supposed to be an architectural prior assisting the processing for relational inference, does not contribute to improved performance here. The SAN model reaches a similar level to the other models only on \dataset{spatial-superlative}, and a slightly worse level of around 91\% on \dataset{spatial-comparative}, while it does not improve on \dataset{spatial-explicit}. Interestingly, even the CNN-LSTM baseline profits from coordinates on \dataset{spatial-superlative}, improving by almost 10\%, despite the fact that positional image embeddings are pooled before being fused with language features.

However, performance of the FiLM model remains virtually unchanged, implying that it does not (solely) rely on coordinates to handle the data. Another aspect which is unique about its architecture is the use of convolutional layers with kernel size $3 \times 3$ as opposed to the fully-connected layers which, in case of the other models, are applied independently per positional embedding. This allows FiLM to capture relative positions locally, and four subsequent such layers are enough to cover the entire $8 \times 8$ feature space (see figure in appendix for full results).

Our first attempt to transfer this insight to the MC model by simply replacing fully-connected layers with convolutions did not improve performance. It turns out that the beneficial effect of convolutions relies on FiLM's feature-wise linear modulation to fuse language and visual features, instead of the otherwise typical concatenation (see figure in appendix for full results). These insights can even be transferred to the CNN-LSTM baseline. Since its late-fusion approach is not able to make full use of spatial information, we also compare to an early-fusion variant where language and visual features are combined at the beginning of its core module (see figure in appendix for full results). Results for MC and CNN-LSTM, included in figure \ref{figure:results2}, confirm the effectiveness of the approach: performance of both models is boosted (almost) to the level of the FiLM model.

Note that integrating this feature is not possible for RelNet, since the pairwise combinations destroy the two-dimensional arrangement of image features which is required to apply convolutions. Similarly, it is not clear how to integrate the changes into SAN's stacked attention layer.

\section{Conclusion}

In this paper we investigate the question: what architectural detail of a VQA model is (or is not) responsible for the ability to correctly handle simple spatial language? By analyzing and comparing various VQA models -- CNN-LSTM, SAN, RelNet, FiLM and MC -- on diagnostic data which specifically targets spatial reasoning capabilities, we identify two alternative techniques whose presence or absence has a deciding impact on whether a model is able to achieve high performance on our data: concatenating image features with relative spatial coordinates (similar to \citep{Liu2018}), or early fusion via feature-wise linear modulation in combination with convolutions. Other features, like the stacked attention layers of SAN or the relation module of RelNet, did not have a beneficial effect.

\subsubsection*{Acknowledgments}

We thank the anonymous reviewers for their constructive feedback. Alexander Kuhnle is grateful for being supported by a Qualcomm Research Studentship and an EPSRC Doctoral Training Studentship.

\bibliography{bibliography}
\bibliographystyle{plain}

\begin{figure*}
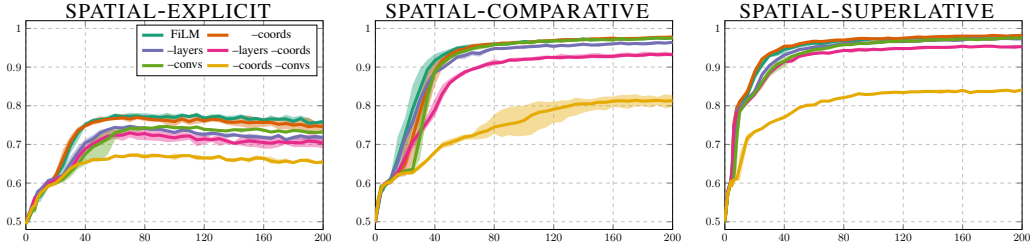

\begin{center}
\textbf{\Large Supplementary material}\\[0.2cm]
to\\[0.1cm]
\textit{What is needed for simple spatial language capabilities in VQA?}
\end{center}
\par\vspace{0.5cm}

\begin{minipage}{0.32\linewidth}
\centering
\dataset{spatial-explicit}\par
\resizebox{\linewidth}{!}{
\begin{tikzpicture}[baseline=(current bounding box.center)]
\begin{axis}[
every axis plot/.append style={line width=2.5pt},
cycle list name=Dark2,
mark options={mark=none},
width=10cm,
height=7.5cm,
xmin=0, xmax=200,
xtick={0,40,80,120,160,200},
xticklabel style={font=\footnotesize},
xmajorgrids,
ymin=0.48, ymax=1.02,
ytick={0.5,0.6,0.7,0.8,0.9,1.0},
ymajorgrids,
yticklabel style={font=\footnotesize},
grid style=dashed,
legend columns=2
]
\input{spatial-explicit-film.tex}
]
\end{axis}
\end{tikzpicture}}
\end{minipage}

\begin{minipage}{0.32\linewidth}
\centering
\dataset{spatial-comparative}\par
\resizebox{\linewidth}{!}{
\begin{tikzpicture}[baseline=(current bounding box.center)]
\begin{axis}[
every axis plot/.append style={line width=2.5pt},
cycle list name=Dark2,
mark options={mark=none},
width=10cm,
height=7.5cm,
xmin=0, xmax=200,
xtick={0,40,80,120,160,200},
xticklabel style={font=\footnotesize},
xmajorgrids,
ymin=0.48, ymax=1.02,
ytick={0.5,0.6,0.7,0.8,0.9,1.0},
ymajorgrids,
yticklabel style={font=\footnotesize},
grid style=dashed,
legend columns=2
]
\input{spatial-comparative-film.tex}
]
\end{axis}
\end{tikzpicture}}
\end{minipage}

\begin{minipage}{0.32\linewidth}
\centering
\dataset{spatial-superlative}\par
\resizebox{\linewidth}{!}{
\begin{tikzpicture}[baseline=(current bounding box.center)]
\begin{axis}[
every axis plot/.append style={line width=2.5pt},
cycle list name=Dark2,
mark options={mark=none},
width=10cm,
height=7.5cm,
xmin=0, xmax=200,
xtick={0,40,80,120,160,200},
xticklabel style={font=\footnotesize},
xmajorgrids,
ymin=0.48, ymax=1.02,
ytick={0.5,0.6,0.7,0.8,0.9,1.0},
ymajorgrids,
yticklabel style={font=\footnotesize},
grid style=dashed,
legend columns=2
]
\input{spatial-superlative-film.tex}
]
\end{axis}
\end{tikzpicture}}
\end{minipage}

\caption{\label{figure:film}%
Accuracy performance curves over the course of training, for the FiLM model and various ablations: no coordinate map (--coords), one as opposed to four FiLM layers (--layers), fully-connected as opposed to convolutional layers (--convs).}
\end{figure*}

\begin{figure*}
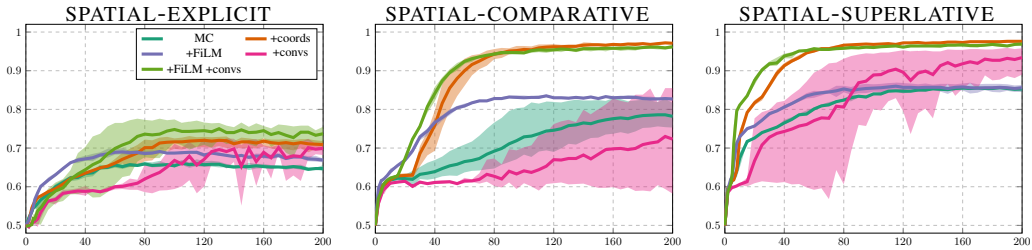


\begin{minipage}{0.32\linewidth}
\centering
\dataset{spatial-explicit}\par
\resizebox{\linewidth}{!}{
\begin{tikzpicture}[baseline=(current bounding box.center)]
\begin{axis}[
every axis plot/.append style={line width=2.5pt},
cycle list name=Dark2,
mark options={mark=none},
width=10cm,
height=7.5cm,
xmin=0, xmax=200,
xtick={0,40,80,120,160,200},
xticklabel style={font=\footnotesize},
xmajorgrids,
ymin=0.48, ymax=1.02,
ytick={0.5,0.6,0.7,0.8,0.9,1.0},
ymajorgrids,
yticklabel style={font=\footnotesize},
grid style=dashed,
legend columns=2
]
\input{spatial-explicit-mc.tex}
]
\end{axis}
\end{tikzpicture}}
\end{minipage}

\begin{minipage}{0.32\linewidth}
\centering
\dataset{spatial-comparative}\par
\resizebox{\linewidth}{!}{
\begin{tikzpicture}[baseline=(current bounding box.center)]
\begin{axis}[
every axis plot/.append style={line width=2.5pt},
cycle list name=Dark2,
mark options={mark=none},
width=10cm,
height=7.5cm,
xmin=0, xmax=200,
xtick={0,40,80,120,160,200},
xticklabel style={font=\footnotesize},
xmajorgrids,
ymin=0.48, ymax=1.02,
ytick={0.5,0.6,0.7,0.8,0.9,1.0},
ymajorgrids,
yticklabel style={font=\footnotesize},
grid style=dashed,
legend columns=2
]
\input{spatial-comparative-mc.tex}
]
\end{axis}
\end{tikzpicture}}
\end{minipage}

\begin{minipage}{0.32\linewidth}
\centering
\dataset{spatial-superlative}\par
\resizebox{\linewidth}{!}{
\begin{tikzpicture}[baseline=(current bounding box.center)]
\begin{axis}[
every axis plot/.append style={line width=2.5pt},
cycle list name=Dark2,
mark options={mark=none},
width=10cm,
height=7.5cm,
xmin=0, xmax=200,
xtick={0,40,80,120,160,200},
xticklabel style={font=\footnotesize},
xmajorgrids,
ymin=0.48, ymax=1.02,
ytick={0.5,0.6,0.7,0.8,0.9,1.0},
ymajorgrids,
yticklabel style={font=\footnotesize},
grid style=dashed,
legend columns=2
]
\input{spatial-superlative-mc.tex}
]
\end{axis}
\end{tikzpicture}}
\end{minipage}

\caption{\label{figure:mc}%
Accuracy performance curves over the course of training, for the MC model and various modifications: with coordinate map (+coords), FiLM fusion as opposed to concatenation (+FiLM), convolutional as opposed to fully-connected layers (+convs).}
\end{figure*}

\begin{figure*}
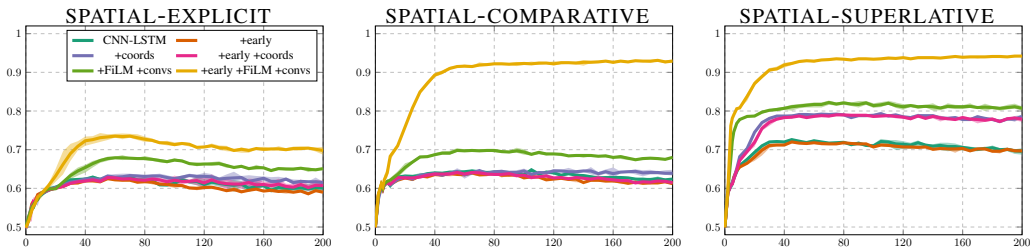


\begin{minipage}{0.32\linewidth}
\centering
\dataset{spatial-explicit}\par
\resizebox{\linewidth}{!}{
\begin{tikzpicture}[baseline=(current bounding box.center)]
\begin{axis}[
every axis plot/.append style={line width=2.5pt},
cycle list name=Dark2,
mark options={mark=none},
width=10cm,
height=7.5cm,
xmin=0, xmax=200,
xtick={0,40,80,120,160,200},
xticklabel style={font=\footnotesize},
xmajorgrids,
ymin=0.48, ymax=1.02,
ytick={0.5,0.6,0.7,0.8,0.9,1.0},
ymajorgrids,
yticklabel style={font=\footnotesize},
grid style=dashed,
legend columns=2
]
\input{spatial-explicit-cnnlstm.tex}
]
\end{axis}
\end{tikzpicture}}
\end{minipage}

\begin{minipage}{0.32\linewidth}
\centering
\dataset{spatial-comparative}\par
\resizebox{\linewidth}{!}{
\begin{tikzpicture}[baseline=(current bounding box.center)]
\begin{axis}[
every axis plot/.append style={line width=2.5pt},
cycle list name=Dark2,
mark options={mark=none},
width=10cm,
height=7.5cm,
xmin=0, xmax=200,
xtick={0,40,80,120,160,200},
xticklabel style={font=\footnotesize},
xmajorgrids,
ymin=0.48, ymax=1.02,
ytick={0.5,0.6,0.7,0.8,0.9,1.0},
ymajorgrids,
yticklabel style={font=\footnotesize},
grid style=dashed,
legend columns=2
]
\input{spatial-comparative-cnnlstm.tex}
]
\end{axis}
\end{tikzpicture}}
\end{minipage}

\begin{minipage}{0.32\linewidth}
\centering
\dataset{spatial-superlative}\par
\resizebox{\linewidth}{!}{
\begin{tikzpicture}[baseline=(current bounding box.center)]
\begin{axis}[
every axis plot/.append style={line width=2.5pt},
cycle list name=Dark2,
mark options={mark=none},
width=10cm,
height=7.5cm,
xmin=0, xmax=200,
xtick={0,40,80,120,160,200},
xticklabel style={font=\footnotesize},
xmajorgrids,
ymin=0.48, ymax=1.02,
ytick={0.5,0.6,0.7,0.8,0.9,1.0},
ymajorgrids,
yticklabel style={font=\footnotesize},
grid style=dashed,
legend columns=2
]
\input{spatial-superlative-cnnlstm.tex}
]
\end{axis}
\end{tikzpicture}}
\end{minipage}

\caption{\label{figure:cnnlstm}%
Accuracy performance curves over the course of training, for the CNN-LSTM model and various modifications: with early fusion (+early), with coordinate map (+coords), FiLM fusion as opposed to concatenation (+FiLM), convolutional layer (+convs).}
\end{figure*}

\begin{figure*}
\vspace*{5cm}
\end{figure*}

\end{document}